\documentclass{article}
\usepackage{spconf,amsmath,graphicx}
\usepackage{amssymb}
\title{An Inter-observer consistent deep adversarial training for visual scanpath prediction}
\name{Author(s) Name(s) \thanks{Thanks to TIC-ART Regional Project for funding this research.}}
\address{Author Affiliation(s)}

\name{Mohamed Amine Kerkouri $^{1*}$\thanks{* The first two authors have equal contributions.}\thanks{This reseach  was funded by the regional TIC-ART Project from Centre-Val de Loire.}, Marouane Tliba$^{1*}$, Aladine Chetouani$^{1}$ and Alessandro Bruno$^{2}$}

\address{$^{1}$Laboratoire PRISME, Université d'Orléans, Orléans, France\\
$^{2}$IULM University , Milan, Italy
\\}
\begin{document}
%\ninept
%
\maketitle
\begin{abstract}
The visual scanpath represents the fundamental concept upon which visual attention research is based. As a result, the ability to predict them has emerged as a crucial task in recent years. It is represented as a sequence of points through which the human gaze moves while exploring a scene. In this paper, we propose an inter-observer consistent adversarial training approach for scanpath prediction through a lightweight deep neural network. The proposed method employs a discriminative neural network as a dynamic loss that better models the natural stochastic phenomenon while maintaining consistency between the distributions related to the subjective nature of scanpaths traversed by different observers. The competitiveness of our approach against state-of-the-art methods is shown through a testing phase.  
\end{abstract}


\vspace{-3mm}
\section{Introduction}
\label{sec:intro}

The human retina receives around $10^{10} bits/sec$ of visual information. Most of this information represents high-definition receptors located in the fovea, which covers approximately $1^\circ$ of the visual field. This gigantic amount of information is further reduced to $3\times10^6 bits/sec$ before traveling through the optical nerve, and further reduced afterward while traveling through the visual cortex \cite{itti2005neurobiology}. The mechanism called "Visual attention" is ruled by the previously mentioned anatomical constraints and other further neurological and psychological ones. The observer is induced to only pay attention to some scene's specific regions. This phenomenon is manifested through saccadic eye movements, representing the gaze shifting from one region to another for a visual stimulus. As eye movements focus on an area, the gaze fixates on specific points, namely "fixation points". The latter can be collected with eye trackers, allowing the projection of fixation points from multiple observers onto a binary map, better known as a "fixation map". On top of that, a "saliency map" is generally obtained with smoothing filters to give a blob-shaped spatial distribution of fixation points over visual stimuli. Saliency maps are usually represented as normalized heatmaps, with each pixel value representing the probability of the pixel catching viewers' attention. The above-depicted mechanism provides the human visual system with outstanding efficiency. The prediction of scanpaths/saliency map is helpful to a lot of computer vision applications like indoor localization \cite{VCIP14}, quality assessment \cite{Jia_Sal, Abouelaziz20NCAA, Aladine18ICIP}%\cite{tliba2022representation}%\cite{tliba2022deep}
, image watermarking \cite{info19Hamidi}, image compression \cite{saliencyComp1}, perception \cite{9715053}, % \cite{SATSal},
and retrieval \cite{retrivalSal}, CVD detection \cite{CVDEyeTracking} . 

The interest of the scientific community in saliency \cite{satsal,salypath360} and scanpath prediction has risen lately. For instance, the winner-take-all (WTA) principle was used by Itti et al. \cite{Itti} in their first work, where the scanpath is extracted from the most salient regions. In \cite{lemeur}, the authors generated scanpaths from a saliency map through statistical features derived from several datasets. In \cite{pathgan}, LSTM layers and the VGG model were employed with adversarial training. In \cite{G-Eymol}, the saliency map was modeled as a gravity field where the gaze mass travels using physical laws. A foveated saliency map was used along with inhibition of return maps to predict scanpaths in \cite{DCSM}. The authors of \cite{SALYPATH} presented an end-to-end model to simultaneously predict the scanpath and the saliency map of an image \cite{SALYPATH}, later generalized for $360^\circ$ images \cite{salypath360}.

The authors of \cite{SSL_Ker} proposed a  self-supervised training approach to train the model for painting image scanpath prediction. While in  \cite{Domain_SP}  they used a domain adaptation approach to generalize the predictive ability from natural scenes to paintings.  

\begin{figure*}[ht!]
  \centering
  \centerline{\includegraphics[scale=0.35 ]{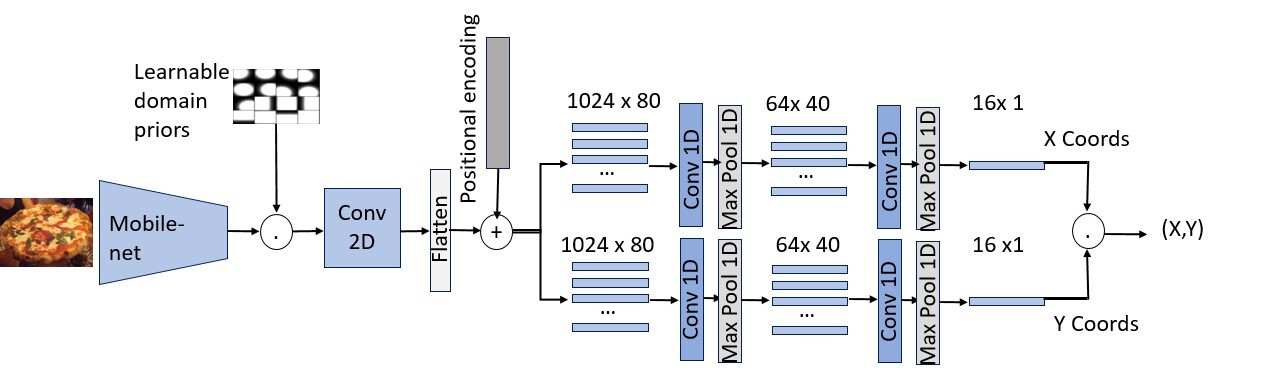}}

\caption{Generator model architecture. }
\label{fig:gen_arch}
\vspace{-6mm}
\end{figure*}

Through these previous works, we found that this task still presents some fundamental and interesting challenges. The stochastic nature of scanpaths is a function of the subjectivity of observers, and modeling this inter-observer distribution in a consistent manner proves to not be an evident task. At the same time, modeling multiple observers induces difficulty for neural networks in generating results that emulate the qualitative properties of the real data. So, the main concern manifests itself in how to train a neural network to predict scanpaths while maintaining consistency between the subjectivities of multiple observers.  

%Challenges:
%\begin{enumerate}
%    \item The difficulty to model scanpaths belonging to different subjective distributions in an inter-observer consistent manner. 
    
%    \item The difficulty to generate plausible scanpaths that emulate the qualitative properties of real data. 

%\end{enumerate}

The proposed method presents the following contribution to solving the aforementioned challenges: 
\begin{itemize}
    \item We employ an adversarial training approach with a min-max game. This dynamic method helps better emphasize the complex nature of scanpaths.
    %\item We use a conditional training method to introduce consistency between data collected from multiple users. 
    \item We condition the learning on the probabilistic distribution of all users, forcing the network to distill the subjective properties of observer population.      
    \item We prove the validity and competitiveness of the proposed method through testing of our model on 2 large datasets. 
\end{itemize}
In the rest of this paper, Section \ref{sec:method} describes the proposed method in detail along with training details. In Section \ref{sec:eval}, we present the experimental protocol as well as discuss the quantitative and qualitative results obtained. Section \ref{sec:conclusion} ends the paper with conclusions.

\vspace{-3mm}
\section{Proposed Method}
\label{sec:method}

To solve the challenges mentioned in Section \ref{sec:intro}, we designed an adversarial training architecture with a thought-out fully convolutional generator model and a discriminator model that is used as a dynamic progressive loss. This later refines the predictive ability of the generator during training by improving its own discriminative ability. This section presents the proposed models (i.e. generator and discriminator) and the training strategies applied.         

\subsection{Generator Architecture}
\label{subsec:arch}

The proposed model utilizes lightweight components to predict scanpaths with variable lengths. Fig. \ref{fig:gen_arch} illustrates the overall architecture of our generator model, which takes an input image and generates a scanpath. To encode the input into a different representational space, we use a pre-trained MobileNet network as a lightweight feature extractor.% MobileNet is considered lightweight due to its use of point-wise and depth-wise separable convolutions, which reduce the number of features necessary for convolutions.% and thus the number of operations to perform.
To enhance the representational ability of our model related to our downstream task, we introduce the use of domain-specific priors through a learnable set of spatially Gaussian distributions, which is a generalization of the "Central bias" theory for visual attention \cite{Cbias}. We model these priors using Eqs. \ref{eq:gaussian2d} and \ref{eq:setgaussian2d}, where $\mu_{x,y}$, $\sigma_{x,y}$  and and $S$ represent the mean of the distribution, the standard deviation and the set of Gaussian priors, respectively.

\begin{equation}
G(x,y) = \frac{1}{2\pi \sigma_x \sigma_y} \exp^{-( \frac{(x-\mu_x)^2}{2\sigma_x^2} + \frac{(y-\mu_y)^2}{2\sigma_y^2})}
\label{eq:gaussian2d}
\end{equation}

\begin{equation}
S = \{ G_1(x,y) , G_2(x,y) , ... , G_{16}(x,y) \}
\label{eq:setgaussian2d}
\end{equation}

In this study, we modeled 16 different Gaussian priors, each with two parameters. The information contained in the set $S$ is then integrated with the features resulting from MobileNet through concatenation followed by a 2D convolution. AS we can consider the scanpaths as an ordinal sequence, we added a positional encoding feature and use a 1D convolutional-based architecture to predict the succession of fixations. More precisely, we used 2 branches of 1D convolutions in order to disentangle the representations of the multi-variable sequence (i.e. the two spatial dimensions). %because during development we found that using d a single branch prevents the network from learning.

%We use 2 different and identical branches in order to disentangle the representations of the multi-variable sequence ( i.e. the two spacial dimensions), as during development we found that using only one branch stops the network from learning.    

%Finally, the generator model encodes the features positionally and uses very lightweight networks to predict appropriate scanpaths.\\
%To be specific, the generator model consists of two parallel branches, designed to disentangle the representations of the 2 dimensions. Each branch uses a succession of 1-dimensional convolutions followed by Max Pooling.

\subsection{Discriminator Architecture}
\label{subsec:disc_arch}

\begin{figure*}[ht!]
  \centering  \centerline{\includegraphics[scale=0.32]{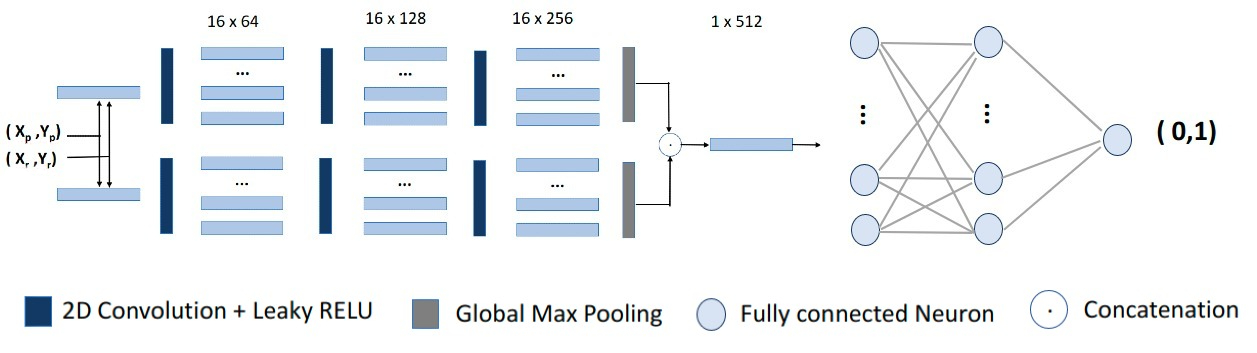}}
\caption{ Discriminator model architecture. }
\label{fig:disc_arch}
\vspace{-4mm}
\end{figure*}

%The second component of the architectural setup is the discriminator network, represented in Fig \ref{fig:disc_arch}. This model aims to discriminate the origin of the input scanpaths by classifying them as being either from the ground-truth distribution or from the fake modeled distribution by the generator model. While training, this model improves its ability to represent the ground truth scanpath distribution. This behaves as a gradually improving and dynamic loss function for the scanpath generator model. We designed the discriminator model to be able to classify the multi-variate sequential data form of the scanpaths. To this end, we separate the sequences representing the coordinates into two different spatial dimensions, allowing the disentanglement of the features of the 2 dimensions. Each branch is composed of a succession of 1D convolutions activated by a Leaky ReLU function with a slope of $0.2$. The feature vectors are gradually increased in proportion to the depth of the network. At the end of each branch, we employ a global Max Pooling Layer on each of the feature vectors. The resulting features from each branch are then integrated through concatenation to construct a global representation of the scanpath. We further employ 3 fully connected layers for the purpose of discriminating the features and classifying the scanpaths. 

The second component of the architectural setup is the discriminator network, illustrated in Fig. \ref{fig:disc_arch}. Its purpose is to discriminate between the distributions of the ground-truth scanpaths and generated ones. During the training, this model enhances its ability to represent the ground-truth scanpath distribution, acting thus as a gradually improving and dynamic loss function for the scanpath generator model.
%We designed the discriminator model to classify the multi-variate sequential data form of the scanpaths. 
Inspired by the generator performance, we separated the sequences representing the coordinates into two different spatial dimensions, enabling the disentanglement of the features of the two dimensions. Each branch of the discriminator model consists of a succession of 1D convolutions activated by a Leaky ReLU function with a slope of $0.2$. The extracted features are gradually increased in proportion to the depth of the network. At the end of each branch, a global Max Pooling Layer is employed on each of the feature vectors. The resulting vectors are then concatenated to build a global representation of the scanpath. Finally, we employed three fully connected layers for discriminating the features and thus classifying the scanpaths.
 
\subsection{Adversarial Training}
\label{subsec:typestyle}

%In order to maintain the consistency of the predicted scanpath with multiple users, we employ a probabilistic conditional adversarial training approach. As scanpaths represent a portion of the complex perceptual feature space, the modeling of such a phenomenon is not an obvious task. Though integrating $NSS$ \cite{NSS} metric with other general machine learning losses \cite{Metrics_scanpath} might introduce a slight improvement. 

In order to model with greater accuracy and maintain consistency of the predicted scanpath with multiple users, we opted to use the min-max game between the classifier discriminator and the predictive generator networks, represented through the following equation: 
\begin{equation}
\begin{split}
\min_{G}\max_{D} V(G,D) = \mathbb{E}_{\hat{y}\sim p_{\text{data}}(y)}[\log{D(\hat{y}|p(y))}] + \\
\mathbb{E}_{x\sim p_{\text{x}}(x)}[1 - \log{D(G(x|p(y))}]
\end{split}
    \label{eq:minmax game}
\end{equation}
where $x$ represents the input image, $p(y)$ is the distribution of different random ground truth scanpaths from multiple users changed periodically during the training, and $\hat{y}$ represents the predicted scanpath. $G$ and $D$ are the generator and discriminator models, respectively. \\

This training approach aims to reduce the distance between the predicted scanpath and the whole set of users, allowing to the integration of the cognitive biases of multiple viewers while maintaining good qualitative shapes for the scanpath. %We found throughout experimentation that normal loss functions do not maintain this last aspect. 
Therefore, this forces the network to learn a non-user-specific representation of the perceptual function. The model was trained for 246 epochs with a learning rate equal to $10^{-5}$.

%Due to the stochastic nature of the scanpaths and in order to preserve the consistency of representation between different users, we regularly change the ground truth scanpath for a specific image on each epoch. This forces the discriminator and, by extension, the generator to learn non-user-specific features during the modeling of scanpaths coming from different subjective origins and distributions. The training lasted for 246 epochs with a $10^{-5}$ learning rate.   

\vspace{-4mm}

\section{Evaluation}
\label{sec:eval}

\subsection{Datasets}
\label{subsec:datasets}
We evaluated our method on two widely-used datasets, namely \textbf{Salicon} \cite{Salicon} and \textbf{MIT1003} \cite{MIT1003}. \textbf{Salicon} \cite{Salicon} consists of 9000 images for training, 1000 images for validation, and 5000 images for testing with the corresponding saliency maps and scanpaths data for all users.

\textbf{MIT1003} \cite{MIT1003} is usually presented along with the MIT300 dataset \cite{MIT300} benchmark. It consists of 1003 natural scene images with the corresponding saliency maps and fixation points gathered throughout eye-tracking sessions. Each image has 15 observers, resulting in 15045 scanpaths.

\subsection{Experimental Protocol}
\label{subsec:Protocol}

In our work, we test on the 5000 images on Salicon dataset with approximately 250000 scanpaths, which ensures the empirical soundness of the results. It is worth noting that in this study our model was trained only on the training set of this dataset. We then used the entire MIT1003 dataset for testing in a cross-dataset evaluation manner without fine-tuning our model on this dataset. Similarly to the first dataset, the significant number of scanpaths ensures empirically sound results. 

To evaluate the performance of our method, we employed three commonly used metrics: MultiMatch, NSS, and Congruency. $MultiMatch$ (MM) metric \cite{Multimatch} compares the similarity of two vectors using five characteristics (Shape, Direction, Length, Position, and Duration). Since the model predicts only spatial coordinates, we use only the first four characteristics and measure the overall performance with their mean value. Two hybrid metrics that compare the predicted scanpaths with a general saliency map for a given image are also employed: $NSS$ and $Congruency$. $NSS$ calculates the mean saliency value of the scanpath fixation locations over the ground truth saliency map, while $Congruency$ computes the ratio of the predicted fixation points which are in the salient regions after thresholding and binarizing the ground truth saliency map. The hybrid metrics allow to measure the accordance and consistency between the predicted scanpath and the users.  

\begin{table*}[htp!]
\begin{center}
\scalebox{0.8}{
\begin{tabular}{ c  c  c  c  c  c  c c}
\hline
\textbf{Model} & \textbf{MM Shape } & \textbf{MM Dir}   & \textbf{MM Len} & \textbf{MM Pos}& \textbf{MM Mean}& \textbf{NSS} & \textbf{Congruency}  \\ 
\hline
 PathGan\cite{pathgan} & 0.9608  &  0.5698   &   0.9530 &     0.8172  &    0.8252 & -0.2904  &  0.0825        \\ 
 \hline
 Le Meur\cite{lemeur} & 0.9505  & 0.6231   &  0.9488 & 0.8605  &   0.8457  &  0.8780 &   0.4784        \\ 
 \hline
 G-Eymol\cite{G-Eymol} & 0.9338 & 0.6271 & 0.9521 & \textbf{0.8967} &  0,8524 & 0.8727 & 0.3449\\
 
 \hline
 SALYPATH  \cite{SALYPATH} & 0.9659  & \textbf{0.6275} & 0.9521 & 0.8965  &   0,8605 &   0.3472 &  0.4572     \\ 
 %\hline
 
 %our model (no Adversarial)  & 0.9552  & \textbf{0.6466} & 0.9509 & 0.8873  &   0.8600 &   \textbf{1.0062} &  0.5170    \\ 
 \hline
 
  our model  (Adversarial)  &  \textbf{0.9745}  & 0.6246 & \textbf{0.9642} & 0.8892  &   \textbf{0.8631} &   \textbf{0.9762} &  \textbf{0.5226}    \\ 
 \hline

\end{tabular}}
\caption{\label{tab:salicon_results}Results of scanpath prediction on Salicon.}
\end{center}
\vspace{-5mm}

\end{table*}

\begin{table*}[htp!]
\begin{center}
\scalebox{0.8}{
\begin{tabular}{ c  c  c  c  c  c  c c}
\hline
\textbf{Model} & \textbf{MM Shape } & \textbf{MM Dir}   & \textbf{MM Len} & \textbf{MM Pos}& \textbf{MM Mean}& \textbf{NSS} & \textbf{Congruency}  \\ 
\hline
 PathGan\cite{pathgan} & 0.9237  & 0.5630   & 0.8929   &  0.8124    &   0.7561  & -0.2750  &  0.0209        \\ 
 \hline
 DCSM (VGG)\cite{DCSM}  & 0.8720 & 0.6420 & 0.8730 & 0.8160 & 0,8007 & - & - \\ % to be seen later 
 \hline
 DCSM (ResNet)\cite{DCSM} & 0.8780 & 0.5890 & 0.8580 & \textbf{0.8220} &  0,7868 & - & -\\ %  to  be seen later
 \hline
 Le Meur\cite{lemeur} & 0.9241  &  0.6378  & 0.9171  &  0.7749 &  0,8135  & 0.8508   &  \textbf{0.1974}    \\ 
 \hline
 G-Eymol\cite{G-Eymol} & 0.8885 & 0.5954 & 0.8580 & 0.7800 &  0,7805 & \textbf{0.8700} & 0.1105\\
 
 \hline
 SALYPATH  \cite{SALYPATH}& 0.9363  & 0.6507 & 0.9046 & 0.7983  &   0,8225 &  0.1595  &   0.0916    \\ 
 %\hline

%our model (no Adversarial)  & 0.9201  & \textbf{0.6759} & 0.9099 & \textbf{0.8351} &   \textbf{0.8352} &   0.8186 &  0.1926    \\ 
 \hline
 
 our model  & \textbf{0.9614}  & \textbf{0.6529} & \textbf{0.9423} & 0.7862 &   \textbf{0.8357} &   0.7523 &  0.1797    \\ 
 \hline
\end{tabular}}
\caption{\label{tab:mit1003_results}Cross dataset evaluation: Results of scanpath prediction on MIT1003.}
\end{center}
\vspace{-4mm}
\end{table*}

\begin{figure*}[ht!]
  \centering
  \centerline{\includegraphics[ scale = 0.39 ]{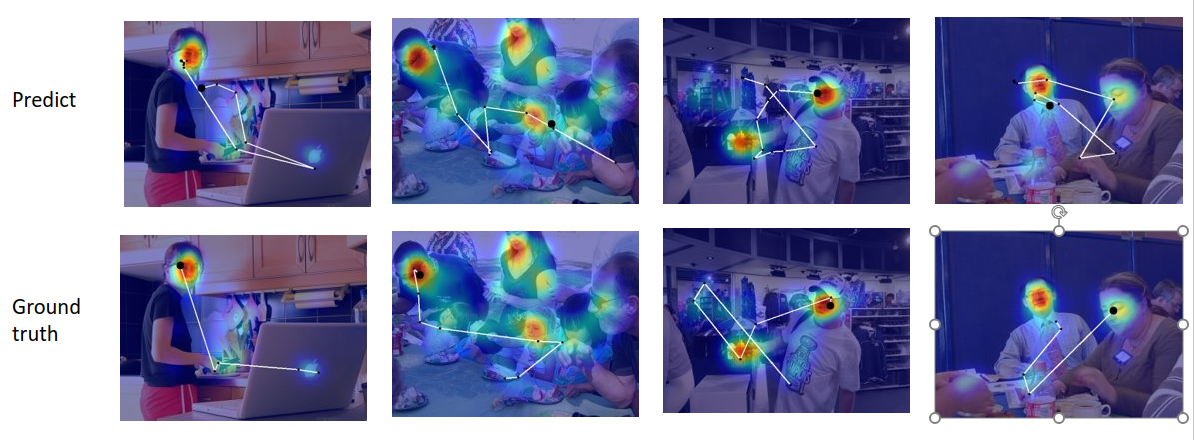}}
\caption{Visualisation results. }
\label{fig:vis}
\vspace{-4mm}
\end{figure*}

\subsection{Quantitative Results}
\label{subsec:Quant}

Tables \ref{tab:salicon_results} and \ref{tab:mit1003_results} show the results obtained after testing our model according  to the protocol described in Section \ref{subsec:Protocol} on the Salicon and MIT1003 datasets, respectively. The performance reached by our method is compared to state-of-the-art methods. 

The results achieved on Salicon (see Table \ref{tab:salicon_results}) show that our model outperformed state-of-the-art methods on the  shape and length components of the $multimatch$ metric. More precisely, we notice a significant improvement in the shape and length components, while the overall mean $multimatch$ shows an improvement compared with the other models. We also achieved the top results for the $congruency$ metric, which indicates that predicted fixations are mostly located in salient regions, maintaining thus a certain consistency with the distribution of users on the Salicon dataset. This is further emphasized and supported by the state-of-the-art results obtained on other metrics. 

As we tested the model on the MIT1003\cite{MIT1003} dataset without any kind of fine-tuning, the results obtained in Table \ref{tab:mit1003_results} show a natural decrease compared to those obtained on Salicon. Nonetheless, the results achieved are still quite high and competitive with the state-of-the-art. This shows the generalization ability of our approach to data distributions coming from different sources. This is especially true, knowing that some of the comparative models trained on the MIT1003 dataset like DCSM \cite{DCSM} and Le Meur\cite{lemeur}. The results show a large improvement in the shape and length components of $multimatch$ compared to other models, while maintaining competitive results for direction. The overall result is competitive compared to the other models. The results on the hybrid metrics (i.e. $NSS$ and $congruency$) maintain close margins with the state-of-the-art since our model was not train on any subset of MIT1003, which has a different distribution  of observers. We can also notice that Le Meur\cite{lemeur} and G-Eymol\cite{G-Eymol} models were able to maintain a slightly better performance because they rely on a saliency map generation step before sampling the scanpath.    

\subsection{Qualitative results}
\label{subsec:Qualt}
Fig. \ref{fig:vis} depicts some qualitative results of predicted scanpaths (i.e. in the center) compared to ground truth scanpaths (i.e. on both sides). %It is clear that predicted scanpaths follow the general allure of the real ones. The majority of the predicted fixation points have the same spatial distribution as the actual ones. 
Our model's predictions show high fidelity to the original scanpaths while maintaining consistency between scanpaths originating from different users, making each predicted scanpath highly plausible. 

\vspace{-3mm}

\section{Conclusion}
\label{sec:conclusion}
\vspace{-4mm}
%The modeling of scanpaths is a complex task on multiple levels, thus finding a suitable training loss is of out-most importance. 
In this paper, we introduced an adversarial training method that uses a discriminative network as a dynamic loss for gradually improving the representative ability of our model, while maintaining inter-observer consistency originating from the subjective nature of scanpaths. We tested our model on the two most used datasets for visual attention modeling and achieved outstanding competitive results on several hybrid and vector-based metrics. The qualitative results showed that our method succeeded to emulate scanpath obtained in the real world. This confirms that substituting traditional loss functions with adversarial training methods would yield better results for complex tasks of perception and attention.      

\vspace{-3mm}
% References should be produced using the bibtex program from suitable
% BiBTeX files (here: strings, refs, manuals). The IEEEbib.bst bibliography
% style file from IEEE produces unsorted bibliography list.
% -------------------------------------------------------------------------
\small
\bibliographystyle{IEEEbib}
\bibliography{strings,refs}

\end{document}